\documentclass{article}
\usepackage{amsfonts}
\usepackage{amsmath}

\newtheorem{theorem}{Theorem}

\newtheorem{notation}[theorem]{Notation}

\input{tcilatex}

\begin{document}

\title{Notes on Geometric Measure Theory Applications to Image Processing;
De-noising, Segmentation, Pattern, Texture, Lines, Gestalt and Occlusion.}
\author{Simon P Morgan \\
University of Minnesota, morgan@math.umn.edu}
\maketitle

\begin{abstract}
Regularization functionals that lower level set boundary length when used
with L$^{1}$ fidelity functionals on signal de-noising on images create
artifacts. These are (i) rounding of corners, (ii) shrinking of radii, (iii)
shrinking of cusps, and (iv) non-smoothing of staircasing. Regularity
functionals based upon total curvature of level set boundaries do not create
artifacts (i) and (ii). An adjusted fidelity term based on the flat norm on
the current (a distributional graph) representing the density of curvature
of level sets boundaries can minimize (iii) by weighting the position of a
cusp. A regularity term to eliminate staircasing can be based upon the mass
of the current representing the graph of an image function or its second
derivatives. Densities on the Grassmann bundle of the Grassmann bundle of
the ambient space of the graph can be used to identify patterns, textures,
occlusion and lines.
\end{abstract}

\tableofcontents

\section{Introduction}

This is an informal discussion paper suggesting possible use of ideas in
geometric measure theory from minimal surfaces, rather than BV, for image
processing. In particular we investigate the use of rectifiable currents in
various spaces to represent signals and images with the flat norm, the
natural topology for currents. We use oriented Grassmann bundles as they are
the spaces that represent both direction and position. As this is an
exploratory discussion paper there are no proofs, references in image
processing, literature review or reports on computational results at this
stage. It is intended that these should be added to subsequent versions of
this paper, along with pictures and where errors and omissions will
hopefully be corrected.

\begin{notation}
$A\subset R^{2}$ is the domain of the image.

$A\times \lbrack 0,1]$

$G_{A}=A\times S^{1}$ denotes the oriented Grassmann bundle of $A.$ This
implicitly involves a choice of basis.

$S^{2}$ will be used to denote the oriented Grassmannian of $\mathbb{R}^{3}$

$S^{4}$ will be used to denote the oriented Grassmannian of $\mathbb{R}^{5}$

$[f]$ is the\ rectifiable current represent the graph of $f$

$\partial \lbrack f]$is the rectifiable current representing boundary of
graph of $f$ which will include discontinuities.

$M(T)$ is the mass of current $T$.

$\mathcal{F}(T)$ is the current flat norm of $T$.
\end{notation}

\section{Examples and philosophy}

\begin{tabular}{|c|c|c|}
\hline
& Mass & Curvature \\ \hline
\begin{tabular}{c}
Regularity \\ 
for \\ 
de-noising \\ 
and \\ 
segmentation \\ 
\end{tabular}
& 
\begin{tabular}{c}
$M([f])+M(\partial \lbrack f])$ \\ 
$(R_{4})$ \\ 
prevents \\ 
staircasing \\ 
on $[f]$and $\partial \lbrack f]$ \\ 
\end{tabular}
& 
\begin{tabular}{c}
Unsigned \\ 
$R_{1},R_{2},R_{3},R_{5}$ \\ 
prevents \\ 
staircasing, no \\ 
shrinking radii \\ 
or smoothing%
\end{tabular}
\\ \hline
\begin{tabular}{c}
Fidelity \\ 
for \\ 
de-noising \\ 
and \\ 
segmentation \\ 
\end{tabular}
& 
\begin{tabular}{c}
$L^{1}(F_{1})$ \\ 
prevents \\ 
average \\ 
values \\ 
drifting \\ 
\end{tabular}
& 
\begin{tabular}{c}
Flat norm \\ 
on signed \\ 
curvature \\ 
prevents \\ 
cusp \\ 
drift%
\end{tabular}
\\ \hline
\begin{tabular}{c}
Pattern \\ 
Texture \\ 
Lines \\ 
Gestalt \\ 
Occlusion \\ 
\end{tabular}
& 
\begin{tabular}{c}
Graph \\ 
density \\ 
and spatial \\ 
frequencies \\ 
\\ 
\end{tabular}
& 
\begin{tabular}{c}
Densities \\ 
and spatial \\ 
frequencies \\ 
on \\ 
$A\times S^{2}$ and \\ 
$A\times \lbrack 0,1]\times S^{2}\times S^{4}$%
\end{tabular}
\\ \hline
\end{tabular}

The aim is to remove noise from a signal $g$ by finding a de-noised image $f$%
. Noise is small scale and randomly structured thus having high curvature.
We can approach this with regularity terms that assign a cost to
irregularity, and a fidelity term which assigns a cost to deviation from a
signal.

We want the regularity terms to be scale invariant, so small versions of a
shape cost the same as large versions, but we want fidelity to not be scale
invariant so we allow small versions to be eliminated for less than large
versions, for the same gain in regularity.

For the regularity we choose total curvature for its scale invariance. For
fidelity we chose the flat norm on curvature for it gets proportionately
cheaper as scale becomes smaller.

An obvious question is why not just use the standard $L^{1}$ norm for
fidelity. This will work fine where level sets are always smooth, but will
introduce distortion on cusps. Therefore we use curvature on the fidelity
term with the flat norm.

\subsection{Large bumps (object) vs small bumps (noise)}

The flat norm on any 1-current corresponding to a bump (level set reduced
boundary curvature density) will reduce more than linearly under homothetic
shrinking.

$F(\lambda _{\frac{1}{c}\#}T)\leq \frac{1}{c}F(T)$

Equality for all $c$ only occurs in special geometric situations with
straight line segments. Therefore noise will follow the strict inequality as 
$t<1$ in the following:

$F(\lambda _{\frac{1}{c}\#}T)=F(T)\{\frac{1}{c}t+(1-t)\frac{1}{c^{2}}\}<%
\frac{1}{c}F(T)$

Now we use flat norms for fidelity terms but not for regularity. This means
removal of small objects cheaper than removing large ones for the same gain
in regularity.

\subsection{Fine undulations (noise) vs coarse undulations (objects)}

Regularity should be unsigned to prevent cancellation, making undulations
expensive, whereas signed measure for fidelity enables cancelling on flat
norm.

\subsection{Line drawings}

In line drawings the level set boundaries are very close and can cancel as
their orientations as currents are opposite. However for low curvatures
there is a low regularity term associated with the curves. There will be
critical line thickness if we represent it as level sets. If however we
represent it as a 2 current of the graph, then we can consider the density
weighting to be by Gaussian curvature. This will give low density to
cylindrical regions or product like regions which is what the graph of a
line is like.

Another possibility is to switch to representing a thin line as a 1 current,
rather than a thick region whose level set boundaries are 1 currents. A way
to detect such lines is to see that the level set boundaries cancel almost
completely under the flat norm but do not under the mass norm, and the mass
is concentrated in certain directions.

\subsubsection{Gestalt and Occlusion}

By considering lines or level set boundaries as 1 currents we are naturally
considering measures in the oriented Grassmann bundle of $A$, $G_{A},$ that
is $A\times S^{1}.$ Consider a projection $p$ of this space onto $L\times
S^{1},$ where $P$ projects $A$\ orthogonally onto $L$. We can allow the
measure in $G_{A}$ to be pushed forward under $p$. The measure in points in $%
S^{1},$ above point $y$ in $L$, corresponding to the directions
perpendicular to $L$ in $A$, will indicate whether or not there was any
measure on the line perpendicular to $L$ passing through $y$. Now we can
detect mass on lines in $A$.

The flat norm can also be used for this purpose. Lift the 1 currents in $A$
to $G_{A}$. Place a cost penalty on boundary and isolated direction change
of lines in $A$. Isolated direction changes in $A$ correspond to currents in 
$G_{A}$, moving in the directions of the local basis of $S^{1}.$ We can
simply increase the current density if it moves in those directions. This
has the practical effect of filling in broken straight lines that correspond
to occlusion or to continuation in patterns. Consider a road intersection
from above. The road is one tone and the sides another. Eight straight line
segments meet in pairs at 90 degrees. Under the flat norm on the currents in 
$G_{A},$ the colinear lines will be filled in. They do not interact or
interfere with each other. There is no need to decide if one road is
occluding the other.

\subsection{Texture and pattern}

Using the projection above for occlusion, we can identify parallel or nearly
lines in patterns and texture. This can be done by taking the projection of
the mass of the currents in $G_{A},$ and determining which directions have
the mass, and what the spatial frequency of the mass is along lines such as $%
L$.

Doing this on bundles such as $A\times \lbrack 0,1]\times S^{2}\times S^{5},$
the oriented Grassmann bundle of the oriented Grassmann bundle, will give
more information.

\section{Level Set Total Curvature Regularization terms}

The type of curvature here is the curvature of boundaries of levels sets as
they are easy to define in terms of functions on domains. Instead it is
possible to consider the curvature of the graph as an embedded submanifold.
This is mentioned below in another section.

We are trying to optimize by varying $f$ for a given $g,$ the energy:

$E(f,g)=\gamma _{1}R_{1}(f)+\gamma _{2}R_{2}(f)+\gamma _{3}R_{3}(f)+\gamma
_{4}R_{4}(f)+\gamma _{5}R_{5}(f)+\gamma _{6}F_{1}(f,g)+\gamma _{7}F_{2}(f,g)$

\subsubsection{The $C^{2}$ term ($R_{1}$)}

Say we have an a.e. $C^{2}$ candidate function $f$ for de-noising. We can
define its total curvature as.

$R_{1}(f)=\int \left\vert \nabla f\right\vert \left\vert \left( \frac{\nabla
f}{\left\vert \nabla f\right\vert }\right) _{J\frac{\nabla f}{\left\vert
\nabla f\right\vert }}\right\vert dxdy$

Here $J\frac{\nabla f}{\left\vert \nabla f\right\vert }$ is a unit vector
perpendicular to $\nabla f$ in the direction given by rotation by $\pi /2.$

\subsection{The discontinuous case ($R_{2}$)}

Say we now allow $f$ to have a 1- dimensional rectifiable set upon which it
is discontinuous. as a graph. Also now we can define the 2-dimensional
rectifiable current $[g]$ as the current representing the set $%
F=\{(x,y,g(x,y))\}$. Now we can define $\partial \lbrack g]$ as the one
dimensional rectifiable current representing $\partial \{(x,y,g(x,y))\}$ the
boundary of $F$ if it is embedded surface in $\mathbb{R}^{3}.$ There may be
isolated points where the graph is not continuous, but these and any set of $%
\mathcal{H}^{2}$ measure zero will be ignored by the currents.

$F_{s}=p\#(Supp\partial \lbrack g])$ where $p$ is projection onto $A$

$R_{2}(f)=\underset{F_{s}}{\int }\left\vert \Delta \theta \right\vert dH^{0}+%
\underset{F_{s}}{\int }\left\vert \theta ^{\prime }\right\vert dH^{1}$

The first term corresponds to angles in the discontinuity, the second to
smooth curves. $\theta $ is the angle of the tangential direction to $F_{s}$

\subsubsection{Generating a discontinuous graph from a signal}

If the gradient of the signal is high and does not vary much over a more or
less convex region of the domain, then the regularity may increase greatly
by introducing a discontinuity. This can be done by choosing a threshold
value for the gradient and then a minimum length and height of
discontinuity. The check that the cut really does reduce regularity costs
without adding to much fidelity cost. Here the choice of unit vector scale
will determine the relative cost of edge length for volume are of $L^{1}$
fidelity.

\subsection{The continuous, non-$C^{1}$ case($R_{3}$)}

There may be polyhedral types of edges on the graph that can be represented
as having curvature. This can given as

$R_{3}(f)=\int \left\vert \Delta \theta \right\vert dH^{1}$

Note that we need $\Delta \theta =0$ almost everywhere.

\section{Fidelity terms ($F_{1},F_{2}$)}

The usual fidelity term given a signal $g$ and candidate de-noised function $%
f$ is

$F_{1}(f,g)=\int \left\vert f-g\right\vert dxdy$

We add a cusp stability term based on the flat norm of 1-currents
representing curvature level set density.

$F_{2}(f,g)$=$\mathcal{F}(F_{\theta }-G_{\theta })$

Where $F_{\theta }=\left[ \left\vert \nabla f\right\vert \left( \frac{\nabla
f}{\left\vert \nabla f\right\vert }\right) _{J\frac{\nabla f}{\left\vert
\nabla f\right\vert }}\right] +[F_{s}]\theta ^{\prime }\left\vert
f_{1}-f_{2}\right\vert +[S]\Delta \theta \left\vert f_{1}-f_{2}\right\vert $
\ (think of lift to $\mathbb{R}^{2}XS^{1}$, but unlifted it is still a
current, just not rectifiable)

Here $[Fs]$ and $[S]$ are 1-currents.

As one currents $F_{\theta }-G_{\theta }=R+\partial T$ where $T$ is a 2
current

$\mathcal{F}(F_{\theta }-G_{\theta })=\underset{R,T}{min}\left(
M(R)+M(T)\right) $ where $F_{\theta }-G_{\theta }=R+\partial T$

This is hard to compute and involves a transport problem.

An approximate way to compute the flat norm is to take $\underset{\phi }{sup}%
(F_{\theta }-G_{\theta })(\phi )$ where $\left\vert \phi \right\vert \leq 1,$
and $\left\Vert \phi \right\Vert =\left\vert d\phi \right\vert \leq 1.$This
will need to be done under a suitable scaling.

\subsection{Transportation cost.}

The second term curvature measure fidelity is evaluated in terms of the flat
norm. Note that is signed. This means that undulations on a small scale will
cancel, and this will give a huge benefit to regularity. However if we have
cusps that create large fidelity terms then they will not move too far
without a big fidelity penalty.

\section{Anti-staircasing regularity ($R_{4},R_{5}$)}

$R_{4}(f)=M([f])+M(\partial \lbrack f])$ as a regularity term will remove
staircasing but may re-introduce distortion. A better approach may be to
take components of curvature in vertical directions i.e. $f_{xx}$, $f_{yy}$, 
$f_{xy}$.Call this $R_{5}=\int \left\vert f_{xx}\right\vert ^{2}+\left\vert
f_{yy}\right\vert ^{2}+\left\vert 2f_{xy}\right\vert ^{2}dxdy$

\section{Scaling and the flat norm vs regularity.}

Here we see that how the length of unit vector is assigned to a signal
domain will determine the scale amount of denoising. This works because the
flat norm works across two different dimensional measures. The length of the
unit vector determines the relationship between these measures.

\subsection{Uniform packing of circles}

Say we have simulated noise of the form of $n^{2}$ discs of radius $1/n$,
and the discs are $2n$ apart. As $n$ varies, this acts like a homothety on a
pattern. If we take as fidelity the 1-current $\theta ^{\prime }\left[ B%
\right] $\ representing the curvature weighted boundary $\left[ B\right] $
of these discs we find the flat norm $\mathcal{F}(\theta ^{\prime }\left[ B%
\right] )$ is

$n^{2}min(2\pi ,\frac{\pi }{n^{2}})=\pi $ \ \ for $n>\frac{1}{\sqrt{2}}$

However if we take absolute value of curvature weighted boundary as
regularity cost we get $n^{2}2\pi $

Clearly eliminating the noise on a scale below where $\pi <n^{2}2\pi $ will
reduce overall energy (Regularity-fidelity).

So for $n>\frac{1}{\sqrt{2}}$, this will occur.

As a result the scale on which the length of unit is set is the order of
size where denoising will eliminate signal features of this form.

\subsection{Edge oscillations}

Say we have a saw tooth with angle $\theta $ from the mean boundary
direction. Let $\frac{2}{n}$ be the period of the sawtooth.

Fidelity is $n.min(2\theta ,\frac{\theta }{n\cos \theta })=\frac{\theta }{%
\cos \theta }$ for large $n....n>1/2cos\theta $

Regularity is $n\theta $

So for $n>1/cos\theta ,$ the saw tooth will be wiped out in optimization.
Again how the unit is scaled will determine how much smoothing occurs.

\subsection{Thin ellipses}

As 1-currents these will self cancel in the flat norm. We can model this
with semicircles of radius $1/n$.

The regularity is $\pi ,$ the flat norm is $\frac{\pi }{2n^{2}}$+$\frac{2}{n}%
.$ So again the scaling of the unit vector in the domain determines the
scale of the de-noising.

\section{Representation and computation}

\subsection{Representing a known current}

Given a 1 current $S$ in R$^{2},$ we have the following representation $%
\{(x,y,a,b):x,y\in n/m$ , $S$ is given by $a(x,y)dx+b(x,y)dy\}$

Similarly a 2 current $T$ in $\mathbb{R}^{3}$ can be represented by $%
\{(x,y,z,a,b,c)x,y,z\in n/m$ where $T$ is given by $%
a(x,y)dydz+b(x,y)dzdx+c(x,y)dxdy\}$

To find a discrete set of values for a given current:

(i) find a polyhedral approximation of the support set by straight line
segments or triangles.

(ii) for each segment or triangle find the average density and the direction
vector coefficients $(a,b)$ or $(a,b,c)$

(iii) assign position coords for the midpoint of the segment or triangle and
combine with direction coeffs giving $(x,y,a,b)$ or $(x,y,z,a,b,c)$

\subsection{Building a current from signal data}

Say we have a level set we wish to represent as a 1-current.

We need positions of the level set and directions, but we also need density
that we are going to take as $\theta ^{\prime }.$ If all the directions are
represented as unit vectors $\mathbf{T}$, and we move along the path of the
boundary at unit speed, $\frac{d\mathbf{T}}{dt}^{{}}$ is $\frac{d\theta
^{\prime }}{dl}.$

\subsection{Optimizing energy of the represented current}

Now when it comes to computing gradient descent of overall cost functional
we need to identify those parts of $S$\ or $T$\ that have a high regularity
cost for a relatively low flat norm penalty.

Regularity cost on a region is given by $\sum \sqrt{a^{2}+b^{2}}$

Flat norm fidelity penalty on a region is given by $\underset{\theta }{max}%
\left( \cos \left( \theta \right) \sum a+\sin \left( \theta \right) \sum
b\right) .$By this we mean what is the contribution to $E(f,g)$ if $%
f=constant$ locally instead of having the local features of $g$ in the
region being summed over.

The flat norm of the difference between an $f$ and $g$ if $f$ and $g$ vary
by diffeomorphism on the domain $A$, is that it is the sum over all features
of the minimum of twice the current mass associated to the feature or the
current mass times the distance moved.

The scale of de-noising determines the size of the region to be taken for
the comparison.

If fidelity penalty is to low, then start moving to reduce regularity cost
by some method

Possible methods

\qquad 1 reduce length of boundary

\qquad 2 on a semi global scale reduce curvature possibly by using $\theta $
information above.

\section{Geometric Measure Theory Books}

Introductions are given in:

Morgan: Frank Morgan; \textit{Geometric measure theory: a beginner's
guide.3rd Ed.(2000) }This is really a guide and contains many pictures,
examples, brief statements of results and can really help readers who are
working through a deeper text.

Hardt,R. and Simon, L. \textit{Seminar on Geometric Measure Theory}.
Birkhauser (1986)

\bigskip

\bigskip A deeper treatment is given in:

LY: Lin and Yang: \textit{Geometric measure theory: an introduction. 1st Ed.
International Press. Boston. (series in advanced Mathematics Volume 1)}

\textit{Lectures on Geometric Measure Theory} by Leon Simon. available from
the Centre for Mathematical Analysis, Australian National University, Volume
3, 1983

The main reference text for this field is \textit{Geometric Measure Theory}:
Federer. 1969 (reprinted by Springer verlag 1996)

\end{document}